\documentclass[letterpaper]{article} 
\usepackage{aaai25}  
\usepackage{times}  
\usepackage{helvet}  
\usepackage{courier}  
\usepackage[hyphens]{url}  
\usepackage{graphicx} 
\usepackage{natbib}  
\usepackage{caption} 
\usepackage{algpseudocode}
\usepackage{algorithm}
\usepackage{amsmath}
\usepackage{amsfonts}
\usepackage{multirow}
\usepackage{tabularx}
\usepackage{subcaption}
\usepackage{makecell}
\usepackage{algpseudocode}
\usepackage{algorithm}
\usepackage{xspace}
\usepackage{booktabs}

\usepackage{siunitx}        
\usepackage{etoolbox}       

\newcommand{\attack}{{\sc \texttt{FuncEvade}}\xspace}
\newcommand{\defence}{{\sc \texttt{FuncMark}}\xspace}

\DeclareMathOperator*{\argmin}{arg\,min}

\newcommand{\spectral}{{\sc \texttt{Spectral}}\xspace}
\newcommand{\peng}{{\sc \texttt{Peng2021}}\xspace}
\newcommand{\ppeng}{{\sc \texttt{Peng2022}}\xspace}

\newcommand{\wang}{{\sc \texttt{Wang2022}}\xspace}

\newcommand{\zhu}{{\sc \texttt{Deep3DMark}}\xspace}

\makeatletter
\DeclareRobustCommand\onedot{\futurelet\@let@token\@onedot}
\def\@onedot{\ifx\@let@token.\else.\null\fi\xspace}
\def\eg{\emph{e.g}\onedot} 
\def\ie{\emph{i.e}\onedot}

\makeatother

\usepackage{newfloat}
\usepackage{listings}
\DeclareCaptionStyle{ruled}{labelfont=normalfont,labelsep=colon,strut=off} 
\floatstyle{ruled}
\newfloat{listing}{tb}{lst}{}
\floatname{listing}{Listing}
\title{Mesh Watermark Removal Attack and Mitigation: A Novel Perspective of Function Space}
\author{
    Xingyu Zhu\textsuperscript{\rm 1,\rm 2},
    Guanhui Ye\textsuperscript{\rm 1},
    Chengdong Dong\textsuperscript{\rm 2},
    Xiapu Luo\textsuperscript{\rm 2},
    Shiyao Zhang\textsuperscript{\rm 1},
    Xuetao Wei\textsuperscript{\rm 1}\thanks{Corresponding Author.}
}
\affiliations{
    \textsuperscript{\rm 1}Department of Computer Science and Engineering, Southern University of Science and Technology, China \\
    \textsuperscript{\rm 2}Department of Computing, Hong Kong Polytechnic University, Hong Kong\\
    12150086@mail.sustech.edu.cn, 12132370@mail.sustech.edu.cn, chengdong.dong@connect.polyu.hk, csxluo@comp.polyu.edu.hk, zhangsy@sustech.edu.cn, weixt@sustech.edu.cn
}

\usepackage{bibentry}

\begin{document}

\maketitle

\begin{abstract}
Mesh watermark embeds secret messages in 3D meshes and decodes the message from watermarked meshes for ownership verification. Current watermarking methods directly hide secret messages in vertex and face sets of meshes. However, mesh is a discrete representation that uses vertex and face sets to describe a continuous signal, which can be discretized in other discrete representations with different vertex and face sets. This raises the question of whether the watermark can still be verified on the different discrete representations of the watermarked mesh. We conduct this research in an attack-then-defense manner by proposing a novel function space mesh watermark removal attack \attack and then mitigating it through function space mesh watermarking \defence. In detail, \attack generates a different discrete representation of a watermarked mesh by extracting it from the signed distance function of the watermarked mesh. We observe that the generated mesh can evade ALL previous watermarking methods. \defence mitigates \attack by watermarking signed distance function through message-guided deformation. Such deformation can survive isosurfacing and thus be inherited by the extracted meshes for further watermark decoding. Extensive experiments demonstrate that \attack achieves 100\% evasion rate among all previous watermarking methods while achieving only 0.3\% evasion rate on \defence. Besides, our \defence performs similarly on other metrics compared to state-of-the-art mesh watermarking methods.
\end{abstract}

%

\section{Introduction}

\begin{figure}
    \centering
    \includegraphics[width=\linewidth]{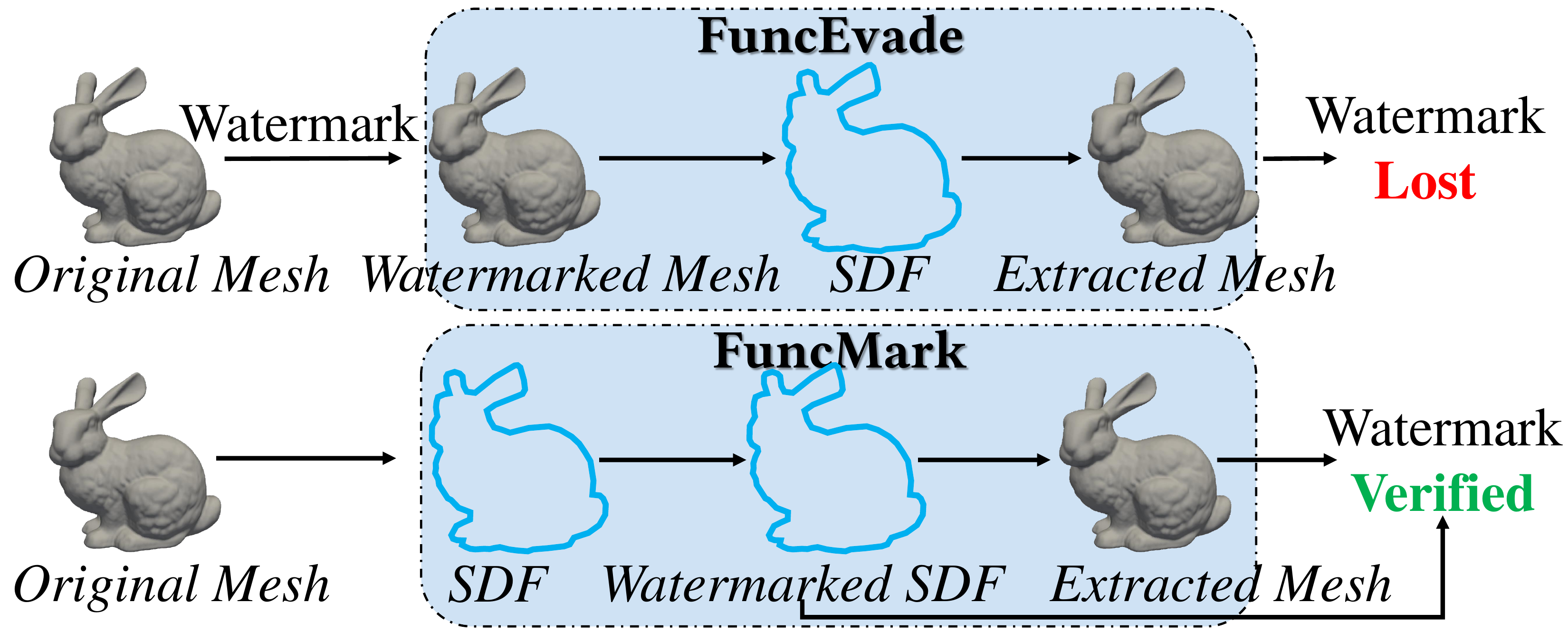}
    \caption{Top: \attack fits the signed distance function (SDF) of the watermarked mesh. It gets an extracted mesh from SDF through isosurfacing, where the extracted mesh successfully evades ALL previous watermarking methods. Bottom: To mitigate \attack, \defence watermark SDF instead of mesh, where the watermark can be verified on SDF and the mesh extracted from it.}
    \label{fig:motivation_example}
\end{figure}



Triangle meshes are the primary representation of 3D geometry in computer graphics. They predominantly represent 3D assets used in video games, movies, manufacturing, and virtual reality interfaces \cite{siddiqui2023meshgpt, pavllo2020convolutional}. Losing a high-fidelity mesh can raise ethical challenges. For example, a mesh stealer can falsely claim ownership of the high-fidelity mesh. One way to address such a challenge is by leveraging digital watermarking techniques \cite{zhu2024rethinking, fernandez2023stable}.

Digital watermarking hides secret messages on digital content like images and meshes for copyright protection, source tracking, and authentication. The watermark is traditionally embedded in the discrete representation of the digital contents. For example, image watermarking embeds a watermark by perturbating image pixels, and mesh watermarking embeds a watermark by perturbing vertex coordinates. However, the underlying signal represented by digital content is continuous. A mesh can be viewed as a discrete representation of a continuous signal, \ie, signed distance function (SDF) \cite{functa, mescheder2019occupancy, siren}, from which a high-quality mesh can be extracted as a different discrete representation. Such a different discrete representation shares the same perceptual quality while having a different vertex and face set (or topology). If we obtain a watermarked mesh, the question is: \textit{can the watermark be verified on its other discrete representations?}

A watermarking method \cite{zhu2024rethinking, wang2022deep, al2019graph, peng2021general, peng2022semi, tsai2022integrating, tsai2020separable, hou2017blind, hou2023separable} consists of embedding, decoding and detection phase. In the embedding phase, the mesh owner embeds a secret message into the original mesh to produce a watermarked mesh. In the decoding phase, a message is decoded from an input mesh (a watermarked mesh or an original mesh without a watermark). In the detection phase, the input mesh is judged as watermarked if the bitwise accuracy of the decoded message is larger than a threshold $\tau$, where bitwise accuracy is the fraction of matched bits in the decoded message and the ground-truth one. For non-learning-based watermarking methods \cite{al2019graph, peng2021general, peng2022semi, tsai2020separable, tsai2022integrating, hou2023separable}, the embedding and decoding algorithm are designed based on heuristics. For learning-based watermarking methods \cite{wang2022deep, zhu2024rethinking}, the embedding and decoding algorithm is built with a deep neural network-based encoder and decoder. Both types of methods embed secret messages on the vertex coordinates of meshes.

Robustness against watermark removal, which post-processes a watermarked mesh to evade watermark detection, is a crucial evaluation metric for watermarking methods. Popular post-processing methods such as Gaussian noise, affine transformation, Laplacian smoothing, and quantization post-process meshes in discrete space, where the post-processed meshes have the same topology and vertex number as the pre-processed meshes, and they only differ in vertex coordinates. \textbf{Research Gap:} existing works \cite{zhu2024rethinking, peng2021general, peng2022semi, wang2022deep} only evaluated the robustness against methods which post-process meshes in discrete space, leaving their robustness against the technique which changes the mesh topology unexplored.

We aim to bridge this gap in an attack-then-defense manner. As shown in Fig. \ref{fig:motivation_example}, we first propose \attack, a function space post-processing method, to evade watermark detection. Given a watermarked mesh, \attack generates its different discrete representation (with different vertex sets and topology) by extracting it from the watermarked mesh's signed distance function (SDF). Compared with discrete space post-process methods, \attack achieves the highest evasion rate while keeping the highest post-processed mesh quality. Our experiment shows that \attack successfully evades all mesh watermarking methods. To mitigate the function space watermark removal, we propose \defence, a function space watermarking method. Instead of watermarking a mesh, \defence watermarks its continuous signal (\ie SDF). \defence embeds secret messages in SDF by spherical partitioning and local deformation. Spherical partitioning divides 3D space into multiple partitions to embed one bit in each partition. The local deformation is performed within each partition based on the bit embedded in the current partition. By doing so, 3D meshes extracted from the deformed SDF inherit such deformation and can be applied for further message decoding. Due to the nature of function space watermarking, \defence successfully mitigates \attack and remeshing attacks. We summarize our contributions as the following:
\begin{itemize}
    \item We take a first step towards investigating the low robustness of all previous mesh watermarking methods under topology attacks, and we propose FuncEvade, a new instance of topology attack that evades watermark detection by generating a different discrete representation of the watermarked mesh.
    \item To mitigate the function space watermark removal, we propose \defence, which embeds a secret message on the signed distance function through spherical partitioning and message-guided deformation. The secret message can be decoded from either the SDF or meshes extracted from the SDF.
    \item We conduct extensive experiments to show that \attack achieves the highest evasion rate among popular removal methods while keeping the highest post-processed mesh quality, and \defence successfully mitigates \attack while keeping similar performance with current SOTA on other metrics.
\end{itemize}
\section{Related Work and Preliminary}

\subsection{Mesh Watermark}

\textbf{Encrypted domain watermarking methods} \cite{jiang2017reversible, tsai2020separable, tsai2022integrating, hou2023separable} encrypt vertex coordinates of a mesh by stream encryption before watermark embedding. The secret message embedding is conducted in the encrypted domain to generate a watermarked encrypted mesh, which can be further used for watermark decoding and plain-text mesh recovery. However, watermark decoding can only be conducted in the encrypted domain because the watermark cannot be verified on the recovered plain-text mesh. The function space of encrypted watermarked mesh is meaningless because encrypted meshes are not meaningful signals. Hence, our work focuses on plain-text domain mesh watermarking.

\textbf{Plain-text domain watermarking methods} can be categorized into learning-based and non-learning-based methods. In non-learning-based methods \cite{al2019graph, peng2021general,peng2022semi, praun1999robustmeshwm}, watermark embedding and decoding are hand-crafted algorithms based on heuristics. In learning-based methods \cite{zhu2024rethinking}, embedding and decoding algorithms use encoders and decoders built with deep neural networks, like Deep3DMark \cite{zhu2024rethinking}. However, most of them only evaluated robustness against watermark removal methods which will not change the topology of the watermarked mesh, leaving the removal methods that will alter the mesh topology (such as remeshing) unexplored. Although \cite{praun1999robustmeshwm} has evaluated their robustness against remeshing, they cannot adapt to current deep learning trends, as they are traditional methods.

\subsection{Signed Distance Function}

The signed distance function (SDF) maps 3D locations $\mathbf{x}$ to a scalar, which represents the signed shortest distance to the 3D surface, with its first derivative representing the surface normal. Given a mesh $\mathcal{M}$ with vertex set $V$, all vertices $v\in V$ satisfy $F(v)=0$, and $\nabla_v F(v)$ equals to the vertex normal at vertex $v$. In this work we follow the tradition \cite{atzmon2020sal, michalkiewicz2019implicit, atzmon2019controlling, chen2019learning, park2019deepsdf,siren, functa} where an SDF is parameterized by a deep neural network $F_\Theta$ with the following unified formulation:
\begin{equation}
    \begin{aligned}
        F_\Theta(\mathbf{x})=\mathbf{W}_n(f_{n-1}\circ f_{n-2}\circ...\circ f_1)(\mathbf{x}),\\
        f_i(\mathbf{x})=\sigma_i(\mathbf{W}_i\mathbf{x}+\mathbf{b}_i),
    \end{aligned}
    \label{eq: SDF_unified}
\end{equation}
where $\mathbf{W}_i, \mathbf{b}_i$ are the weight matrix and bias of the $i$-th layer, and $\sigma_i$ is an element-wise nonlinear activation function. $\sigma_i$ is either ReLU or sinusoidal function used in SIREN \cite{siren}. SIREN pioneeringly applied sine transform to the input coordinates, enabling SDFs to represent high-frequency details better. This work uses SIREN as the backbone for our SDF parameterization.

Given an SDF, a mesh $\mathcal{M}$ can be extracted from it through isosurfacing. Isosurfacing first divides 3D space with voxel cubes with resolution $r^3$, \ie, the 3D space is divided by $r^3$ voxel cubes. Given a scalar function $F(\mathbf{x})$, isosurfacing fits a surface to the points whose sample values are of a specific isovalue and whose position is determined by the edges of voxel cubes. For example, to extract a mesh from a signed distance function is to sample points $\mathbf{x}$ whose scalar value $F(\mathbf{x})=0$. These sampled points are then polygonized to build a surface model through marching cube \cite{lorensen1998MC}.

\begin{table*}
    \centering
    \small
    \begin{tabular}{c| cc|cc|cc|cc|cc}
    \toprule
    \multirow{2}{*}{Removal Methods} & \multicolumn{2}{c|}{\spectral} & \multicolumn{2}{c|}{\peng} & \multicolumn{2}{c|}{\ppeng} & \multicolumn{2}{c|}{\wang} & \multicolumn{2}{c}{\zhu} \\\cline{2-11}
    & EVA & HD & EVA & HD & EVA & HD & EVA & HD & EVA & HD \\\hline\hline

    Gauss Noise & 98.3\% & 0.021 & 97.8\% & 0.021 & 98.2\% & 0.021 & 0.20\% & 0.021 & 0.1\% & 0.021 \\
    Rotation & 80.2\% & / & 92.1\% & / & 0\% & / & 18.4\% & / & 15.2\% & / \\
    Quantization & 94.5\% & 0.015 & 93.8\% & 0.014 & 91.2\% & 0.014 & 0\% & 0.015 & 4.5\% & 0.014 \\
    Smoothing & 99.3\% & 0.108 & 99.7\% & 0.107 & 99.1\% & 0.112 & 14.6\% & 0.116 & 4.9\% & 0.118 \\
    Remesh & \textbf{100\%} & 0.005 & \textbf{100\%} & 0.006 & \textbf{100\%} & 0.005 & \textbf{100\%} & 0.006 & \textbf{100\%} & 0.005 \\
    \attack (Ours) & \textbf{100\%} & \textbf{0.004} & \textbf{100\%} & \textbf{0.003} & \textbf{100\%} & \textbf{0.004} & \textbf{100\%} & \textbf{0.003} & \textbf{100\%} & \textbf{0.004} \\
    \bottomrule
    \end{tabular}
    \caption{Watermark removal results. EVA and HD indicate evasion rate and Hausdorff distance. HD is evaluated between watermarked mesh and post-processed meshes. Lower HD means higher post-processed mesh quality. Note that "/" means that using HD to evaluate rotated mesh similarity is unreasonable because the rotated mesh is the same as the original one.}
    \label{tab:attack_success_rate}
\end{table*}

\section{Mesh Watermark Detection} \label{sec:watermark_detection}
Alice embeds $n$-bit binary message (as her signature) into a mesh. The watermark decoding algorithm then decodes messages from the mesh it receives and detects the watermark when the message is close to the ground truth message. We judge whether the watermark is detected through the following test.

\textbf{Statistical test.} Let $w\in \{0,1\}^n$ be Alice's signature. We extract the message $w'$ from a mesh (the mesh can be non-watermarked mesh $\mathcal{M}$ or watermarked mesh $\mathcal{M}_w$) and compare it to $w$. As done in previous works \cite{luo2023copyrnerf, fernandez2023stable, jiang2023evading}, the detection test relies on the number of bitwise accuracy $BA(w, w')$: if
\begin{equation}
    BA(w, w')\geq \frac{\tau}{n} \quad \text{where}\quad \tau\in \{0,...,n\},
\end{equation}
then the mesh is judged as watermarked. This provides a level of robustness to imperfections of the watermarking.

We test the statistical hypothesis $H_1$: ``The given mesh was watermarked by Alice" against the null hypothesis $H_0$: ``The given mesh was not watermarked by Alice". Under $H_0$, bits $w'_0, ..., w'_{n-1}$ are (i.i.d.) Bernoulli random variables with parameter 0.5 (\ie{} $BA(w,w')\sim B(k, 0.5)/n$). 
The False Positive Rate under $\tau$ (denoted as $FPR(\tau)$) is the probability that $BA(w, w')$ takes a value bigger than the threshold $\frac{\tau}{n}$. Formally, $FPR(\tau)$ can be written as:
\begin{equation}
    \begin{aligned}
        FPR(\tau) &= \text{Pr}(n\cdot BA(w, w')\geq \tau)\\
        &=\sum_{i=\tau}^n \binom{n}{i} \frac{1}{2^n}.
    \end{aligned}
\end{equation}
The key point is to set the threshold $\tau$ such that the $FPR(\tau)$, \ie{}, the probability that an original mesh is falsely detected as watermarked is bounded by a small value $\eta$, \eg{}, $\eta=0.05$. To make $FPR(\tau)<\eta$, $\tau$ should be at least $\tau^*=\argmin_\tau FPR(\tau)<\eta$. For instance, when $n=48$ and $\eta=0.05$, we have $\tau\geq\tau^*=31$.

\section{Watermark Removal Analysis}

We first introduce \attack, which generates a different discrete representation of the watermarked mesh by extracting it from the signed distance function of the watermarked mesh. We compare the removal effectiveness among \attack, remeshing, and other popular watermark removal methods such as Gauss noise, quantization, smoothing, and rotation. Besides, we expose and explain the vulnerability of all previous watermarking methods when facing function space watermark removal methods.

\textbf{Removal Objective.} Given a watermarked mesh $\mathcal{M}_w$, watermark removal aims to get a post-processed mesh $\mathcal{M}'_w$ so that $\mathcal{M}'_w$ can evade watermark detection and $\mathcal{M}'_w$ and $\mathcal{M}_w$ are highly similar. We evaluate evasion effectiveness by \textbf{evasion rate} and mesh similarity by \textbf{Hausdorff distance}. In the watermark removal problem, we expect a higher evasion rate and lower Hausdorff distance.

\subsection{Removal Implementation}\label{sec:removal_impl}

Since the watermark removal method requires high similarity between the watermarked mesh $\mathcal{M}_w$ and post-processed $\mathcal{M}'_w$, and \attack extracts $\mathcal{M}'_w$ from SDF, one challenge is to build a high-quality SDF from the watermarked mesh $\mathcal{M}_w$. SDF predicts the signed shortest distance to the 3D surface given arbitrary 3D points $\mathbf{x}\in \mathbb{R}^3$. Hence, the first step is to sample points $\mathbf{x}$ and their ground truth SDF value to supervise the fitting of SDF.

\textbf{Sampling strategy on mesh.} In practice, sampling strategy significantly impacts the fitted SDF quality. Uniform sampling across $[-1,1]^3$ usually gives sparse on-surface points, thus resulting in low surface fitting quality. We expect on-surface points to be accurately supervised, and off-surface points far from the surface are roughly supervised because we only need to extract zero-isosurface in the mesh extraction stage. We use package \textit{mesh-to-sdf}\footnote{https://github.com/marian42/mesh\_to\_sdf} to produce ground truth SDF for 5M off-surface point set $V_1$ and ground truth normal vectors for 5M on-surface point set $V_0$.

\textbf{SDF fitting strategy.} We supervise SDF fitting process with both its function values $F_\Theta$ and first-order derivatives $\nabla F_\Theta$. Besides, SDF fitting requires solving a particular Eikonal boundary value problem that constrains the norm of first-order derivatives $|\nabla F_\Theta|$ to be 1 everywhere. Hence, we supervise SDF with the following objectives.
\begin{equation}
    \begin{aligned}
        &\mathcal{L}_{F_\Theta}=\int_{\mathbf{x}\in V_0} \lambda_1 (1-\left<\nabla F_\Theta(\mathbf{x}), \nabla F(\mathbf{x})\right >) d\mathbf{x}+\\
        &\int_{\mathbf{x}\in V_0\cup V_1} \lambda_2\lVert F_\Theta(\mathbf{x})-F(\mathbf{x})\rVert+\lambda_3 \lVert |\nabla F_\Theta(\mathbf{x})|-1\rVert d\mathbf{x}\\
    \end{aligned}
    \label{eq:sdf_fitting}
\end{equation}
where $\lambda_1=100, \lambda_2=3000, \lambda_3=5$. For on-surface points $V_0$, their ground truth signed distance values $F(\mathbf{x})$ and first-order derivatives $\nabla F_\Theta(\mathbf{x})$ are assigned with zero and their normal vectors, respectively. For off-surface points $V_1$, their ground truth signed distance values are assigned with values obtained in the sampling process. We use SIREN \cite{siren} as the backbone of $F_\Theta$, which consists of four \textit{Linear} layers with their channel size set to 512. We use Adam optimizer with the initialized learning rate $10^{-3}$ for 1000 epochs, and we decrease the learning rate by half every 200 epochs. After obtaining the fitted SDF, the final step is to obtain the post-processed mesh $\mathcal{M}'_w$ by extracting it from SDF through marching cube \cite{lorensen1998MC}.

\subsection{Removal Evaluation} \label{sec:attac_eva}

\textbf{Metrics and Datasets.} We evaluate the evasion rate for the effectiveness of watermark removal and Hausdorff distance for mesh similarity between the watermarked mesh $\mathcal{M}_w$ and the post-processed mesh $\mathcal{M}'_w$. Let $w'$ be the message extracted from $\mathcal{M}'_w$. We judge the removal method successfully evades the detection if $BA(w, w_A)<\frac{\tau}{n}$. We select the message length $n=48$ and $\tau=31$ in our experiment. In this case, we have a false positive rate $FPR(\tau)<0.05$. We normalize meshes in ShapeNet \cite{shapenet2015} and Stanford Repo \cite{stanfordbunny} to $[-1,1]^3$. All watermark embedding, extraction, and removal experiments are conducted on normalized meshes.

\textbf{Examined watermarking methods.} We examined both learning-based and non-learning-based methods. We denote them as \spectral \cite{al2019graph}, \peng \cite{peng2021general}, \ppeng \cite{peng2022semi}, \wang \cite{wang2022deep} and \zhu \cite{zhu2024rethinking} for brevity. In detail, \spectral decomposes the Laplacian matrix of a mesh into eigenvalues and eigenvectors and hides secret messages in eigenvalues. Both \peng and \ppeng transform vertex coordinates into a new coordinate system before they embed or decode the watermark on the vertex coordinates. The new coordinate system of \peng is built based on the largest face of the input mesh, while \ppeng is built based on the bounding sphere of the input mesh. \wang and \zhu train the encoder and decoder to embed or decode the watermark on the vertex coordinates, where they increase their robustness by adding an attack layer between the encoder and decoder for adversarial training \cite{adversarial_training}. All these methods watermark meshes in vertex coordinates without changing the topology of meshes.

\textbf{Examined post-processed methods.} Following the above watermark process, we post-process watermarked mesh to remove the watermark. We first examine popular mesh processing methods such as Gauss noise, rotation, quantization, and smoothing, which are commonly used in previous works \cite{wang2022deep, zhu2024rethinking} to evaluate the robustness of their watermark. These post-processed methods output $\mathcal{M}'_w$ sharing the same topology with $\mathcal{M}_w$. Our \attack and remeshing \cite{botsch2004remeshing, botsch2010polygon} will output a $\mathcal{M}'_w$ with different topology with $\mathcal{M}_w$. In detail, remeshing incrementally performs simple operations such as edge splits, edge collapses, edge flips, and Laplacian smoothing. All the vertices of the remeshed patch are reprojected to the original surface to keep an approximation of the input. We set the target edge length of remeshing to be half of the average edge length of the input mesh. For other removal methods, we use Gauss noise with mean $\mu=0$ and standard deviation $\sigma=0.005$, rotation with rotation axis $(1,0,0)$ and rotation angle $\alpha<\frac{\pi}{3}$, quantization with bits $N_b=6$, laplacian smoothing with $\lambda=2$ and smoothing iteration $10$.

\noindent \textbf{Question 1.} How is the quality of the mesh post-processed by \attack?

Table \ref{tab:attack_success_rate} uses the previously mentioned removal parameters mentioned. Table \ref{tab:attack_success_rate} shows quantitative results of evasion rate and mesh similarity between $\mathcal{M}_w$ and $\mathcal{M}'_w$. We can observe that (1) Gauss noise, smoothing and quantization make visible distortions to $\mathcal{M}_w$ but cannot evade \zhu and \wang; (2) meshes post-processed by \attack achieves 100\% evasion rate on all watermarking methods while keeping the lowest Hausdorff distance between $\mathcal{M}_w$ and $\mathcal{M}'_w$ (thus highest similarity).




\begin{figure*}[htb]
    \centering
    \includegraphics[width=\linewidth]{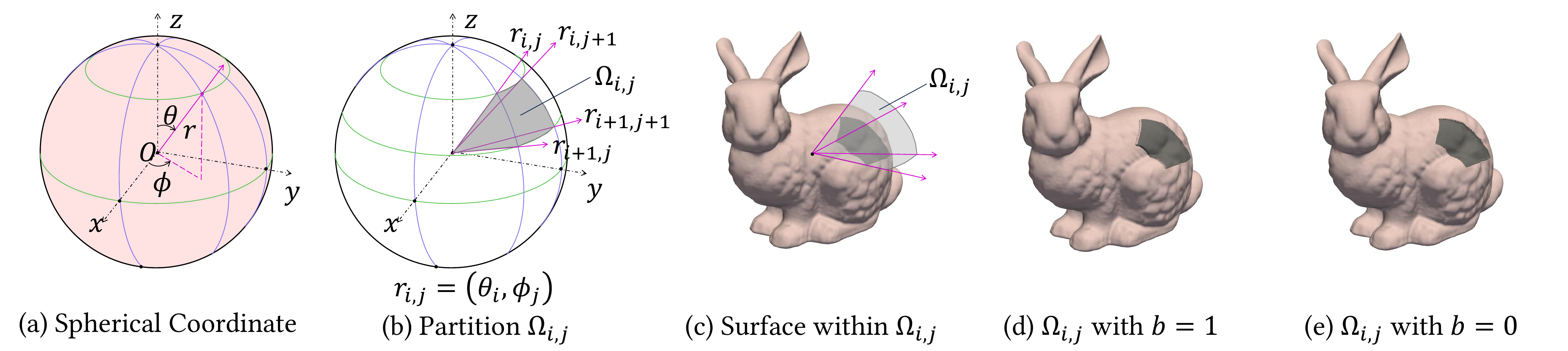}
    \caption{\defence overview. We convert vertex coordinates into (a) spherical coordinate system, which is further divided into $N_s*N_s$ partitions (denoted as $\Omega_{i,j}$). We further embed one bit in each (b) partition $\Omega_{i,j}$. The surface within $\Omega_{i,j}$ is deformed (c) outward if the embedded bit $b=1$, or else deformed (d) inward. The deformation strength is $10\times$ of the default setting.}
    \label{fig:defence_pipeline}
\end{figure*}

\noindent \textbf{Question 2.} Why are previous watermarking methods vulnerable to function space removal?

The critical vulnerability of all previous methods is that their watermark detection is conducted on discretized space. Concretely, a mesh can be viewed as a discretization of a continuous surface. A continuous surface can be discretized in multiple ways. For example, meshes can be extracted from an SDF at different resolutions, resulting in various numbers of vertices and other topologies. However, previous watermarking methods are either vulnerable to topology changes \cite{wang2022deep, zhu2024rethinking, al2019graph, peng2021general} or vulnerable to vertex number and vertex range changes \cite{peng2022semi}. For \spectral, topology changes will result in different Laplacian matrices and thus affect the decomposed eigenvalue where the watermark is hidden. For \wang and \zhu, both methods rely on graph neural networks, which are sensitive to topology changes. For \peng and \ppeng, they transform vertex coordinates into a new coordinate system before embedding or decoding the watermark. The watermark cannot be decoded if the coordinate system changes. \peng builds the coordinate system based on the largest face of the input mesh, which is fragile to topology changes. \ppeng builds the coordinate system based on the bounding sphere and the farthest vertex pair, which is fragile to vertex number and range changes.

If watermark embedding and decoding in discrete space is vulnerable to function space removal, \textit{can we conduct watermark embedding and decoding in function space?}

\section{Function Space Watermarking}

The above analysis finds that current mesh watermarking methods are vulnerable to function space attacks. We address this issue by proposing \defence, which watermarks SDF (mesh in function space) such that the watermark can be verified on the given watermarked SDF or meshes extracted from it. The key idea of \defence is to make local deformations of SDF based on the binary message $w\in\{0,1\}^n$. As Fig. \ref{fig:defence_pipeline} shows, (1) we first divide 3D space and SDF into $N_s*N_s$ partitions under the spherical coordinate system (Fig. \ref{fig:defence_pipeline}(a,b)); (2) within each partition $\Omega_{i,j}$ we embed one bit $w_k\in{0,1}$; (3) the surface within $\Omega_{i,j}$ is either deformed outward if $w_k=1$, or else inward. We compare the effectiveness with previous watermarking methods and evaluate the robustness of \defence against \attack and remeshing.

\textbf{Watermark Objective.} Given a mesh $\mathcal{M}$, watermark aims to get a watermarked mesh $\mathcal{M}_w$ such that $\mathcal{M}_w$ are highly similar to the original mesh $\mathcal{M}$ and the watermark can withstand arbitrary removal attacks. We evaluate robustness by \textbf{evasion rate} and mesh similarity by \textbf{Hausdorff distance}. Unlike the removal problem, we expect lower Hausdorff distance and lower evasion rate in the watermarking problem because a lower evasion rate means higher robustness against watermark removal attacks.

\subsection{Spherical Partitioning} \label{sec: div}

Our first step is to divide 3D space under a spherical coordinate system since we only embed one bit in each partition. A cartesian coordinate $(x,y,z)$ and spherical coordinate $(r,\theta, \phi)$ can be converted to each other by the following:
\begin{equation}
    \begin{split}
        \left \{
        \begin{aligned}
            &r=\sqrt{x^2+y^2+z^2}\\
            &\theta=\arccos(\frac{z}{r})\\
            &\phi=\arctan(\frac{y}{x})
        \end{aligned},
        \right .
        \quad
        \begin{split}
            \left \{
            \begin{aligned}
                &x=r\sin(\theta)\cos(\phi)\\
                &y=r\sin(\theta)\sin(\phi)\\
                &z=r\cos(\theta)
            \end{aligned}
            \right .
        \end{split}.
    \end{split}
    \label{eq: cat_sph}
\end{equation}
We equally divide $\theta$ and $\phi$ into $N_s$ partitions. In this case, the 3D space is divided into $N_s*N_s$ partitions. As Fig. \ref{fig:defence_pipeline}(b) shows, a spherical partition $\mathbf{\Omega}_{i,j}$ is a local region $\{(r,\theta,\phi)|\theta\in[\theta_i, \theta_{i+1}] \cap \phi\in[\phi_j, \phi_{j+1}]]\}$.

\subsection{Watermark Embedding} \label{sec: embed}

\begin{figure*}
    \centering
    \begin{subfigure}[b]{0.24\linewidth}
        \includegraphics[width=\linewidth, page=1]{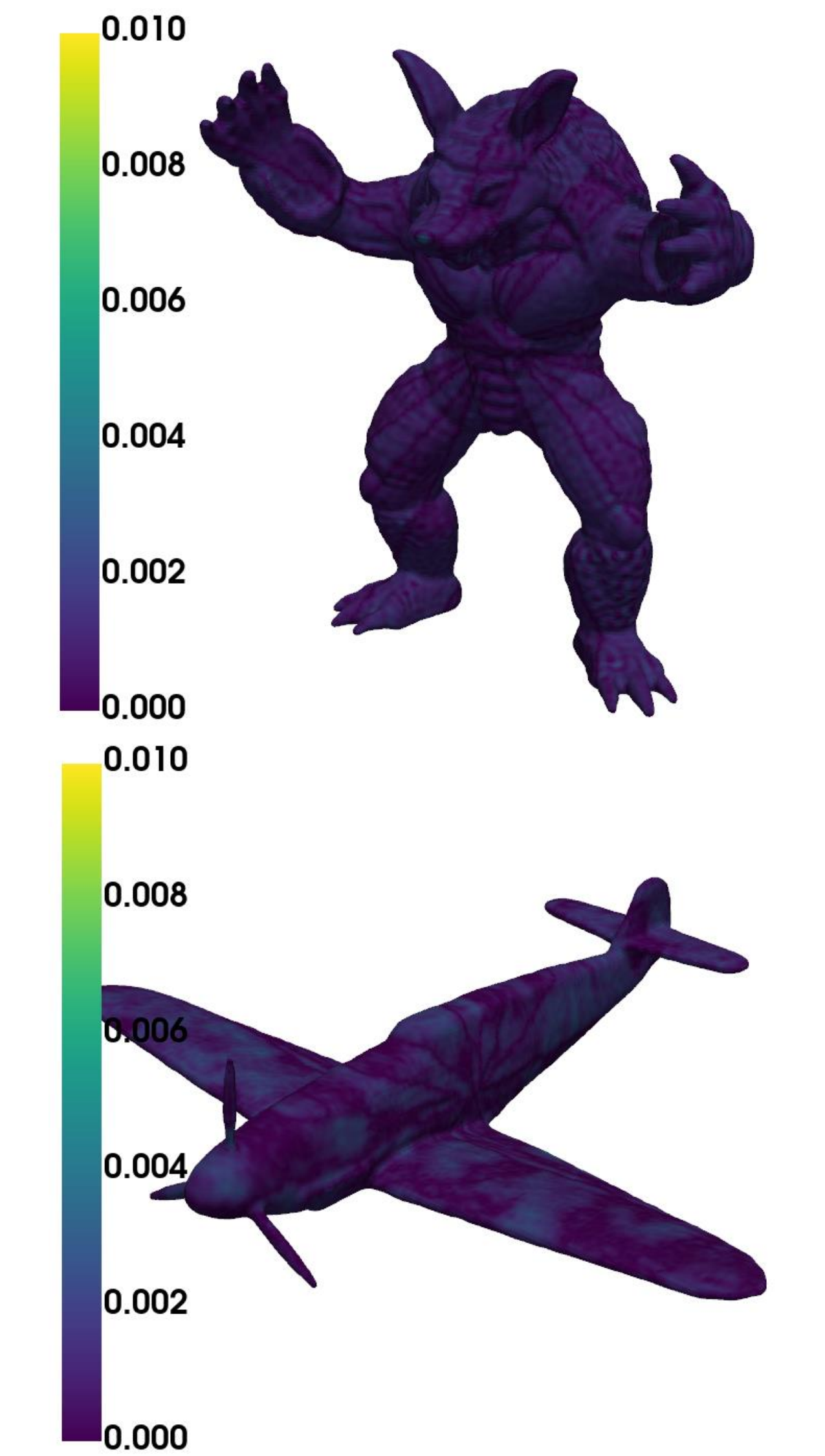}
        \caption{Shortest distance to original mesh for each vertex}
    \end{subfigure}
    \begin{subfigure}[b]{0.24\linewidth}
        \includegraphics[width=\linewidth, page=2]{figures/experiment/visual.pdf}
        \caption{The original mesh}
    \end{subfigure}
    \begin{subfigure}[b]{0.24\linewidth}
        \includegraphics[width=\linewidth, page=3]{figures/experiment/visual.pdf}
        \caption{Mesh extracted from $G_\Theta$}
    \end{subfigure}
    \begin{subfigure}[b]{0.24\linewidth}
        \includegraphics[width=\linewidth, page=4]{figures/experiment/visual.pdf}
        \caption{Visualize vertex tags $b$}
    \end{subfigure}
    \caption{(a-c) Geometric differences between meshes extracted from $F_\Theta$ and $G_\Theta$. (d) Vertex tags $b$ on meshes from $G_\Theta$.}
    \label{fig:visual_mesh}
\end{figure*}

Our watermark is embedded by deforming the SDF $F_\Theta$ based on the embedded bit $w_k$ to get a deformed SDF $G_\Theta$. Suppose the partition $\Omega_{i,j}$ is embedded with bit $w_k$, the surface within partition $\Omega_{i,j}$ will be deformed outward if $w_k=1$, or else inward if $w_k=0$. We define deformation function $\mathbf{x}=D(\mathbf{y})$, which moves the original $\mathbf{y}$ to target point $\mathbf{x}$ as follows: 
\begin{equation}
    D(\mathbf{y})=
    \begin{cases}
        \mathbf{y}+\delta\nabla F_\Theta(\mathbf{y})&w_k=1\\
        \mathbf{y}-\delta\nabla F_\Theta(\mathbf{y})&w_k=0
    \end{cases},
    \label{eq: D}
\end{equation}
where $\delta$ is a constant value representing the deformation strength. The deformed SDF $G_\Theta$ can be represented as $G_\Theta(\mathbf{x})=F_\Theta(D^{-1}(\mathbf{x}))$ \cite{NFGP, deng2021deformed, liu2021editing}. One challenge is to make $D(\mathbf{y})$ invertible. One way to make $D$ invertible is to approximate $D$ with invertible residual blocks \cite{NFGP}. However, in practice, we find the accuracy of invertible residual blocks cannot satisfy the accuracy requirements of watermarking. Fortunately, our $D(\mathbf{y})$ is explicitly defined, which makes it possible for us to derive $D^{-1}(\mathbf{x})$ via Newton's method.


\textbf{Newton's method calculates $D^{-1}(\mathbf{x})$.} We derive $\mathbf{y}=D^{-1}(\mathbf{x})$ by calcuating the zero point of $\mathbf{y}\pm \delta \nabla F_\Theta(\mathbf{y})-\mathbf{x}=0$. We apply Newton's method \cite{newtonmethod} to calculate the zero point as the following repeated process:
\begin{equation}
    \mathbf{y}_{n+1}=\mathbf{y}_{n}-\mathcal{J}^{-1}_{D}(\mathbf{y}_{n})\cdot D(\mathbf{y}_{n}),
    \label{eq: newton_iteration}
\end{equation}
where $\mathbf{y}_{i}$ is the guessed zero point at $i$-th iteration, $\cdot$ is the matrix multiplication operation, and $\mathcal{J}^{-1}_{D}(\mathbf{y})$ is the inverse of Jacobian matrix of $D(\mathbf{y})$. To calculate a correct zero point, this process is sensitive to the accuracy of $\mathcal{J}_{D}(\mathbf{y})$, which is derived from Hessian of the SDF: $\mathcal{H}_{F_\Theta}(\mathbf{y})$. However, $\nabla F_\Theta(\mathbf{y})$ is supervised with ground truth value only when $\mathbf{y}$ is on zero-isosurface, which means $\nabla F_\Theta(\mathbf{y})$ may not be accurate when $\mathbf{y}$ is far from surface. Thus, Newton's method may not find a zero point if $\mathbf{y}_0$ is initialized far from the surface. To address this issue, we randomly initialize multiple $\mathbf{y}_0$ within $[-1,1]^3$. One hundred samples are enough to find the ground truth zero point.

\subsection{Watermark Decoding} \label{sec: decode}

We decode each bit $w_k'$ independently within each partition. A partition $\Omega_{i,j}$ is tagged with bit $b$ if more than half of the points in that region are tagged with bit $b$. A point $\mathbf{x}$ is tagged with $b=1$ if $F_\Theta(\mathbf{x})>0$, otherwise we tag it with bit 0. We can decode watermarks on both $G_\Theta$ and meshes extracted from $G_\Theta$. Given a mesh extracted from $G_\Theta$, we directly tag its vertex set with the binary bit. Given a watermarked SDF $G_\Theta$, we can either extract a mesh from it through isosurfacing or sampling points on its zero-isosurface.

\textbf{Sampling strategy on SDF.} For zero-isosurface sampling, we adopt a sample-and-reject strategy. In detail, we sample points within $[-1,1]^3$ and reject points whose distance to the surface is greater than $10^{-4}$. We then run gradient descent to approximate on-surface points: $\mathbf{x}_{t+1}= \mathbf{x}_{t}-F_\Theta(\mathbf{x}_{t})\nabla F_\Theta(\mathbf{x}_{t})$.

\begin{table*}
    \centering
    \small
    \begin{tabular}{c|c|c|c|c|c|c|c}
    \multirow{2}{*}{Methods} & \multirow{2}{*}{HD$\downarrow$} & \multirow{2}{*}{Accuracy$\uparrow$} & \multicolumn{5}{c}{Evasion Rate$\downarrow$} \\\cline{4-8}
     &  &  & Gaussian & Rotation & Translation & Remesh & \attack (Ours) \\
     \hline\hline
     \spectral & 0.012 & 87.48\% & 100\% & 100\% & 100\% & 100\% & 100\% \\
     \peng & 0.015 & 97.98\% & 100\% & 100\% & 100\% & 100\% & 100\% \\
     \ppeng & 0.037 & 80.89\% & 100\% & \textbf{0\%} & \textbf{0\%} & 100\% & 100\% \\
     \wang & 0.185 & 97.95\% & 0.2\% & 18.4\% & \textbf{0\%} & 100\% & 100\% \\
     \zhu & 0.061 & \textbf{98.17\%} & \textbf{0.1\%} & 15.2\% & \textbf{0\%} & 100\% & 100\% \\
     \defence (Ours) & \textbf{0.004} & 95.69\% & 1.6\% & 14.8\% & \textbf{0\%} & \textbf{0.1\%} & \textbf{0.3\%} \\
    \bottomrule[1pt]
    \end{tabular}
    \caption{Watermark results. HD indicates Hausdorff distance, which is evaluated between the original and watermarked mesh. Accuracy is evaluated before the watermarked mesh is post-processed by removal methods. Lower HD means higher watermarked mesh quality, and a lower evasion rate means higher robustness.}
    \label{tab:traditional}
\end{table*}
\begin{figure*}
    \centering
    \begin{subfigure}[b]{0.33\linewidth}
        \includegraphics[width=\linewidth, page=1]{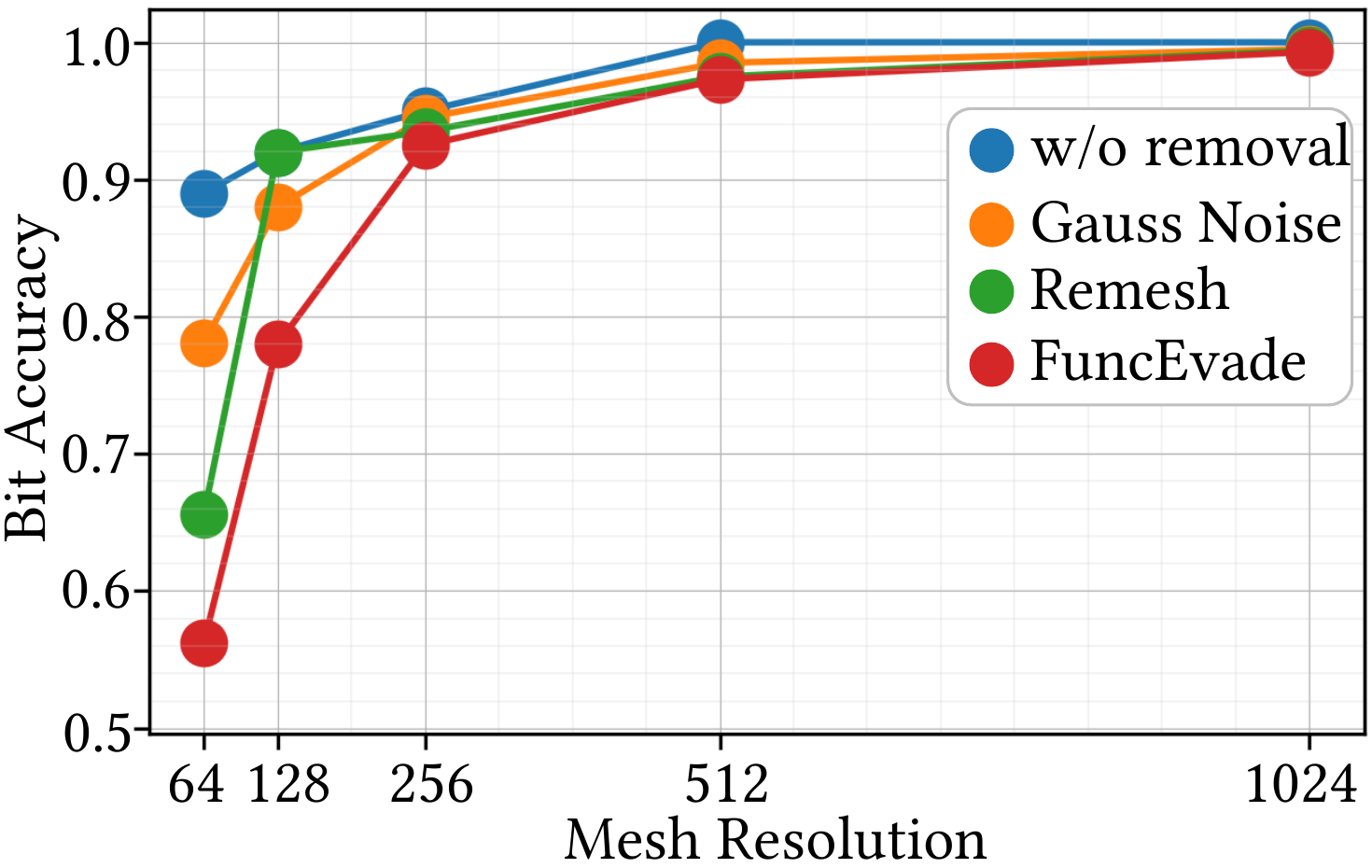}
        \caption{Bit accuracy with varied mesh resolution}
    \end{subfigure}
    \begin{subfigure}[b]{0.33\linewidth}
        \includegraphics[width=\linewidth, page=2]{figures/experiment/plot_nv.pdf}
        \caption{Evasion rate with varied mesh resolution}
    \end{subfigure}
    \begin{subfigure}[b]{0.33\linewidth}
        \includegraphics[width=\linewidth, page=3]{figures/experiment/plot_nv.pdf}
        \caption{Bit accuracy with varied sampled points}
    \end{subfigure}
    \caption{Our watermark can be verified on SDF $G_\Theta$ and extracted meshes. For watermark detection on extracted meshes, we evaluate (a) bit accuracy with varied mesh resolution and (b) evasion rate with varied mesh resolution. For watermark verification on SDF, we evaluate bit accuracy with a varied number of sampled points on $G_\Theta$.}
    \label{fig:plot_nv}
\end{figure*}

\subsection{Watermark Evaluation} \label{sec:defence_eval}

\textbf{Setup.} Since we embed each spherical partition with one binary bit in the secret message $w$, if the number of partitions $N_s*N_s$ is greater than the length of the message, we repeatedly embed message bits until we consume all partitions. Unless explicitly mentioned, our experiment is conducted under the following settings. We set $N_s=32$ (\ie, the spherical system is divided into $32*32$ partitions). We set message length $n=48$, and the detection threshold $\tau=31$. We use marching cube \cite{lorensen1998MC} as our default isosurfacing method and the default watermarking strength $\delta=0.001$. We use the same dataset, evaluation metrics, and watermarking methods (as baselines) in Sec. \ref{sec:attac_eva} to evaluate watermark accuracy before any post-processing method is applied. We use the same set of attack settings for all post-processing methods in Sec. \ref{sec:attac_eva} to compare evasion rates between baselines and \defence.

\noindent \textbf{Question 1.} How is the watermarked mesh quality?

We present visualizations of the shortest distance from vertices of the watermarked mesh to vertices of the original mesh (Figure \ref{fig:visual_mesh}(a)), the original mesh itself (Figure \ref{fig:visual_mesh}(b)), the watermarked mesh extracted from the watermarked SDF $G_\Theta$ (Figure \ref{fig:visual_mesh}(c)), and vertex tag $b$ (Figure \ref{fig:visual_mesh}(d)). Visualizing vertex tags helps better understand the process of embedding the secret message $w$. Notably, it can be observed that the watermarked mesh extracted from the watermarked signed distance function $G_\Theta$ appears perceptually identical to the original mesh.

\noindent \textbf{Question 2.} How is the mesh quality, bit accuracy, and evasion rate compared with other watermarking methods?

Mesh quality is evaluated by Hausdorff distance (HD), where lower HD indicates a higher similarity between the original and the watermarked mesh. Before the removal attack, we evaluate the bit accuracy of watermarking methods to prove they are all effective without attack. Then, we evaluate their robustness against attacks by evasion rate. Table \ref{tab:traditional} shows the performance results between \defence and previous watermarking methods. We can observe that (1) \defence achieves the lowest HD, which means we have the highest mesh quality. (2) Remeshing and \attack have 100\% evasion rate on other watermarking methods while having only 0.1\% and 0.3\% evasion rate on \defence, which means \defence is the only method robust against remeshing and \attack.

\noindent \textbf{Question 3.} Since \defence is a function space watermarking method, can watermark be verified on meshes extracted at arbitrary resolutions?

The watermark of \defence can be verified on both SDF $G_\Theta$ and meshes extracted from $G_\Theta$ at arbitrary resolutions. For watermark decoding on a mesh, where we can directly tag $0/1$ on the mesh vertices, we evaluate the bit accuracy and evasion rate on meshes extracted at arbitrary resolution. For watermark verification on SDF $G_\Theta$, where we have to sample points on the zero-isosurface of SDF such that we can tag $0/1$ on the sampled points, we evaluate the bit accuracy on varied sampled points. Figure \ref{fig:plot_nv} shows the bit accuracy and evasion rate with varied mesh resolution and sampled points. We can observe that (1) all the evasion rates drop to zero on meshes extracted at resolution $256$, and meshes extracted at a resolution lower than $256$ have bad quality, which means \defence mitigates all the removal methods because the attacker cannot evade watermark detection without damaging the resultant mesh quality. (2) Given a watermarked $sdf$ $G_\Theta$, our watermark can be verified with only 250 sampled points.

\section{Conclusion}

In this paper, we have noticed the research gap where no previous work investigated the mesh watermark robustness against function space removal. We have bridged this gap in an attack-then-defense manner by proposing \attack and \defence. \attack is a function space removal method that generates a different discrete representation of the watermarked mesh by extracting it from the signed distance function of the watermarked mesh. We have observed that \attack evades all previous mesh watermarking methods. \defence is a function space mesh watermarking method which successfully mitigates \attack by watermarking signed distance function. By doing so, the watermark can be verified on arbitrary mesh extracted from the watermarked signed distance function. Our experiments have shown that \defence mitigates \attack by dropping its evasion rate from 100\% to 0.3\% while keeping similar performance on other metrics compared with current SOTA methods.

\section*{Acknowledgments}
This work was supported in part by National Key R\&D Program of China under Grant 2021YFF0900300, in part by Guangdong Key Program under Grant 2021QN02X166, and in part by Research Institute of Trustworthy Autonomous Systems under Grant C211153201. Any opinions, findings, and conclusions or recommendations expressed in this material are those of the author(s) and do not necessarily reflect the views of the funding parties.

\bibliography{aaai25}

\end{document}